\pdfoutput=1
\documentclass[11pt]{article}

\usepackage[final]{acl}
\usepackage{amsmath,amssymb}
\usepackage{tcolorbox}

\usepackage{times}
\usepackage{latexsym}
\usepackage{multirow}
\usepackage[T1]{fontenc}
\usepackage[utf8]{inputenc}
\usepackage{booktabs}
\usepackage{makecell}
\usepackage{microtype}

\usepackage{inconsolata}

\usepackage{graphicx}

\title{MolGround: A Benchmark for Molecular Grounding}

\author{
 \textbf{Jiaxin Wu\textsuperscript{1}},
 \textbf{Ting Zhang\textsuperscript{1}},
 \textbf{Rubing Chen\textsuperscript{1}},
 \textbf{Wengyu Zhang\textsuperscript{1}},
\\
 \textbf{Chen Jason Zhang\textsuperscript{1}},
 \textbf{Xiao-Yong Wei\textsuperscript{2,1,}\thanks{Corresponding Author}},
 \textbf{Li Qing\textsuperscript{1}},
\\
 \textsuperscript{1}The Hong Kong Polytechnic University,
 \textsuperscript{2}Sichuan University,
}

\begin{document}

\maketitle

\begin{abstract}

Current molecular understanding approaches predominantly focus on the descriptive aspect of human perception, providing broad, topic-level insights. However, the referential aspect—linking molecular concepts to specific structural components—remains largely unexplored. 
To address this gap, we propose a molecular grounding benchmark designed to evaluate a model's referential abilities.  
We align molecular grounding with established conventions in NLP, cheminformatics, and molecular science, showcasing the potential of NLP techniques to advance molecular understanding within the AI for Science movement. 
Furthermore, we constructed the largest molecular understanding benchmark to date, comprising 117k QA pairs, and developed a multi-agent grounding prototype as proof of concept. 
This system outperforms existing models, including GPT-4o, and its grounding outputs have been integrated to enhance traditional tasks such as molecular captioning and ATC (Anatomical, Therapeutic, Chemical) classification. 
%
% The source code will be released at \textcolor{magenta}{\url{https://github.com/open_upon_acceptance}}.
\vspace{-0.05in}
\end{abstract}

\section{Introduction}
Deep learning models have transformed traditional molecular understanding tasks, including property prediction \cite{ moleculeNet,Walters_deep_learning_property_prediction,zhang2024pre}, molecular generation \cite{deep_learning_molecular_generation, hua2024mudiff,song2024equivariant}, and reaction prediction \cite{Deep_learning_reaction_prediction, ding2024exploring,tavakoli2024ai}. 
%
% These models typically encode molecular structures (e.g., SMILES strings, molecular graphs) into global embeddings, which represent the entire molecule as a single vector for label prediction.  
%
Recently, tasks like molecular captioning \cite{ChEBI-20text2mol} and molecule-language translation \cite{ACLMoleculeCaptionWorkshop} have gained significant attention due to advancements in large language models \cite{molReGPT2024TKDE, bioT52024ACL}.  
These models represent molecular structures as sequences of tokens, enabling the generation of natural language descriptions by leveraging sophisticated sequence-to-sequence learning techniques.  

While having yielded promising results, these approaches primarily mimic the \textbf{descriptive} aspect of human perception ~\cite{cocchiarella1974fregean,geach1950russell,Kamp1993DRT}, focusing on broad, topic-level understanding. 
The \textbf{referential} aspect of perception, which associates concepts with specific molecular components (e.g., atoms, functional groups, rings), has been overlooked.  
For example, consider the SMILES \textit{CC(=O)O} (\textit{acetic acid}). 
In molecular captioning, a typical output might be: ``This is acetic acid, commonly known as the main component of vinegar. It is used industrially in production and exhibits toxic effects at high concentrations.''  
While this description is highly informative, it is descriptive in nature. From a referential perspective, it is more critical to identify which specific part of the molecule contributes to its toxicity.  
In this case, the \textit{carbonyl group} \textit{(C=O) }is responsible for the molecule's corrosive effects, as it facilitates the release of protons \textit{(H+)}, which can damage biological tissues.
This referential understanding not only enhances interpretability but also generalizes to other similar compounds, such as \textit{formic acid}, \textit{oxalic acid}, and \textit{trichloroacetic acid}.  

% This referential understanding not only enhances interpretability but also generalizes to other similar compounds, such as \textit{formic acid} \textit{C(=O)O}, \textit{oxalic acid} \textit{C(=O)(O)C(=O)O}, and \textit{trichloroacetic acid} \textit{C(Cl)(Cl)(Cl)(=O)O}.  

% By emphasizing referential associations, models could provide deeper insights that are both interpretable and applicable across related compounds.

\begin{figure*}[]  
\centering   
  \centering  
  \includegraphics[width=1\linewidth]{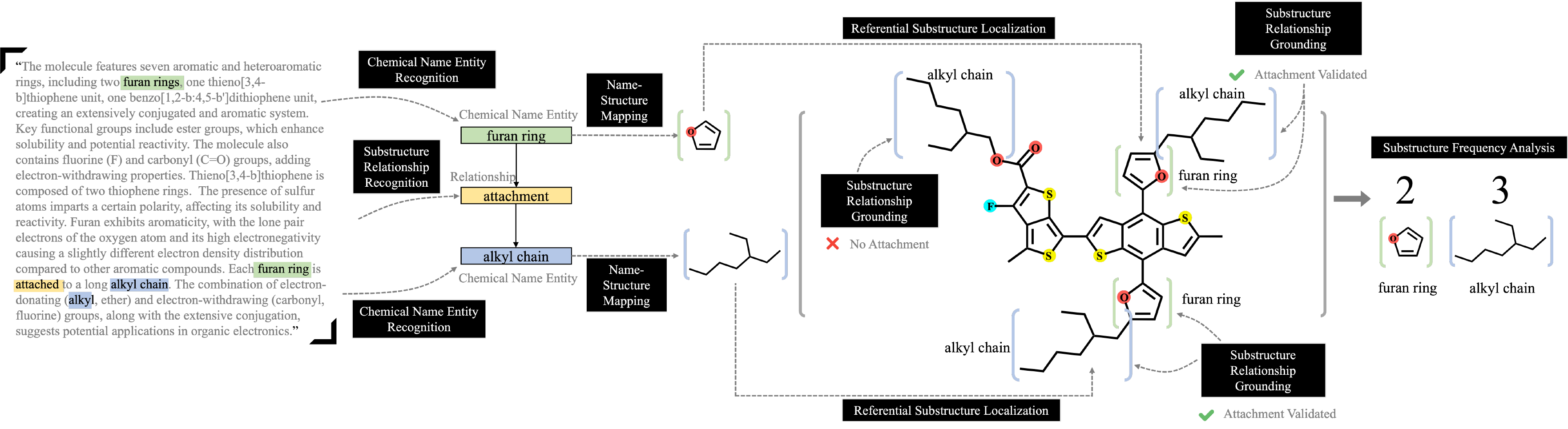}  
    \caption{A referential framework for fine-grained molecular grounding, comprising five tasks: Chemical Name Entity Recognition, Name-Structure Mapping, Referential Substructure Localization, Substructure Relationship Grounding, and Substructure Frequency Analysis, demonstrated through a running example.}   
    \label{fig:groundingProcess}
    % \vspace{-0.25in}
\end{figure*} 

While the complementary nature of descriptive and referential perceptions has long been modeled in cognitive science, such as in Fregean Semantics~\cite{cocchiarella1974fregean}, Russell's Theory of Descriptions~\cite{geach1950russell}, and Discourse Representation Theory (DRT) by Hans Kamp and Uwe Reyle \cite{Kamp1993DRT}, it has also been successfully implemented recently in vision-language research~\cite{arai2024large,liu2023llava}.  
The integration of visual grounding~\cite{xiao2024towards, Deng_2021_ICCV}, which mimics referential perception by linking textual concepts to specific image regions, has significantly advanced the performance, interpretability, and generalization of vision-language models. 
These models, which traditionally relied solely on image-caption pairs for training, have greatly benefited from this approach.  

Believing that molecular understanding research is at a similar turning point, we propose a grounding benchmark to assess a model's ability to explicitly associate molecular concepts with specific structural components.
This benchmark emphasizes fine-grained understanding and interpretability, enabling models to identify, explain, and reason about the roles of particular molecular features.
Unlike visual grounding, where a model is primarily tasked with identifying the locations of concepts, molecular grounding requires the identification of specified components at multiple cognitive levels, including concept instances, structural locations, and compositional facts.  
From a pragmatics perspective, molecular grounding differs from existing molecular understanding tasks that focus on the topic. 
Instead, it emphasizes providing answers to fine-grained queries such as ``What?'', ``Where?'', and ``Which ones?''.  
Figure~\ref{fig:groundingProcess} illustrates the proposed molecular grounding tasks including Chemical Name Entity Recognition (CNER), Name-Structure Mapping (NSM), Referential Substructure Localization (RSL), Substructure Relationship Grounding (SRG), and Substructure Frequency Analysis (SFA). 
This paper serves as a pilot study aimed at formulating molecular grounding by aligning it with established conventions in NLP, cheminformatics, and molecular science.
Our findings demonstrate that NLP techniques can play a critical role in advancing molecular understanding within the broader AI for Science movement.
In addition to creating the largest molecular understanding benchmark to date, we developed a multi-agent grounding prototype as a proof of concept. 
This system outperforms existing models, such as GPT-4o \cite{GPT4o2024openai}, and its grounding results have been successfully integrated to enhance conventional tasks like molecular captioning 
% 
% and molecular classification.
and ATC (anatomical, therapeutic, chemical) classification.

\section{Related Work}
% overview of current tasks in related to descriptive and referential, highlight the complementary nature

Molecular understanding has been a long-standing field of research, predating the recent surge of interest in AI for Science.  
The tasks in this field can be broadly grouped into three categories based on their popularity: 1) Property prediction~\cite{moleculeNet, Walters_deep_learning_property_prediction, Zhang2024ChemLLMAC} and representation learning~\cite{molecular_representation_learning, zhang2024pre}, which are the most extensively studied and widely popular. 
2) Structure prediction~\cite{alphafold, song2024equivariant}, captioning~\cite{molReGPT2024TKDE, ChEBI-20text2mol}, and generation~\cite{deep_learning_molecular_generation, hua2024mudiff}, which have recently gained significant attention. 
3) Emerging studies on tasks such as reaction prediction and optimization~\cite{Deep_learning_reaction_prediction}, interaction prediction~\cite{tavakoli2024ai}, simulations and dynamics~\cite{vander2024atlas}, toxicity and safety assessment~\cite{sahu2024removal}, and visualization and explainability~\cite{janissen2024single}.
The popularity of the first two groups largely stems from the ease of directly applying sophisticated machine learning models to these tasks. 
Early approaches relied on methods like Bayesian classifiers~\cite{langley1992analysis}, logistic regression~\cite{hosmer2013applied}, and SVMs~\cite{hearst1998support}, while more recent efforts have widely adopted CNNs~\cite{o2015introduction}, GNNs~\cite{wu2020comprehensive}, and Transformer-based models~\cite{vaswani2017attention}.
Most of these models follow a pipeline of encoding molecules into embeddings and predicting outputs such as labels or textual descriptions, reflecting the way these models were initially designed.  
However, these implementations tend to be \textbf{descriptive}, as they focus on high-level concepts by treating a molecule as a whole, rather than addressing its internal components.  
The third group, on the other hand, signals a shift toward more fine-grained modeling and improved interpretability. 
This shift is driven by two factors: 1) The needs of identifying subcomponents in molecular science, such as reaction tracing \cite{smith2007substructure2rections, Li_substructure_reaction_tracing, Umit_retrosynthetic_prediction} and understanding molecule-target interactions~\cite{positionDrugDesign1997, segler_planning_2018}.
2) Advancements in machine learning for interpretability and generalization~\cite{gao2023interpretability}.  
Ultimately, these developments highlight the growing demand for models with \textbf{referential} perception, enabling them to go beyond high-level descriptions and address specific components within a molecule.

\vspace{-0.03in}
% theoretical background of congnitive science and lessons learned from vision-language research 

The complementary relationship between descriptive and referential perceptions has been extensively explored in cognitive science, as seen in Fregean Semantics (senses and references)~\cite{cocchiarella1974fregean}, Russell's Theory of Descriptions (definite descriptions and proper names)~\cite{geach1950russell}, and Discourse Representation Theory (descriptions and referents)~\cite{Kamp1993DRT}.
However, referential perception has not been explicitly modeled or systematically evaluated in molecular understanding.
From this perspective, Table~\ref{Tab:descriptiveVSreferential} summarizes advanced models such as BioT5 \cite{bioT52024ACL}, ChemLLM \cite{Zhang2024ChemLLMAC}, and Mol-Instructions \cite{fang2024molinstructions}, alongside commonly adopted benchmarks like ChemBench4K \cite{Zhang2024ChemLLMAC}, and MoleculeQA \cite{lu-etal-2024-moleculeqa}.
It is clear that this is an area requiring more focused and explicit attention.
While recent advancements in models have made strides toward incorporating referential perception, with promising results observed in integrating referential perception-oriented visual grounding (e.g., RefFormer~\cite{wang2024referencingfocusimprovingvisualgrounding}, ClawMachine~\cite{ma2024clawmachine}, DOrA~\cite{wu2024dora}), significant challenges remain. 
This results from the heavy reliance on costly human expertise for benchmark construction and the lack of a systematic formulation of the problem.
As highlighted in Table~\ref{Tab:descriptiveVSreferential} , our proposed MolGround represents an initial effort to address these challenges. 
It scales up to 1.91 times larger than existing benchmarks and introduces fine-grained definitions to better align with the requirements of referential perception.

% Please add the following required packages to your document preamble:
% \usepackage{multirow}
\begin{table}[]
    \resizebox{1\linewidth}{!}{
\begin{tabular}{c|l|r|r|r}
\toprule
      \textbf{Benchmarks}               & \textbf{Tasks}                                       & \textbf{\#QA} & \textbf{\#Des.(\%)} &\textbf{\#Ref.(\%)} \\ \hline
\multirow{10}{*}{CB4} & Caption2mol                                 & 800                      & 97.75                           & 2.25                            \\
                              & Mol2Caption                                 & 299                      & 100.00                          & 0.00                            \\
                              & Name Conversion                             & 799                      & 99.87                           & 0.13                            \\
                              & Product Prediction                          & 300                      & 96.99                           & 3.01                            \\
                              & Yield Prediction                            & 300                      & 100.00                          & 0.00                            \\
                              & Temperature Prediction                      & 202                      & 98.98                           & 1.02                            \\
                              & Solvent Prediction                          & 300                      & 87.66                           & 12.34                           \\
                              & Retrosynthesis                              & 300                      & 96.00                           & 4.00                            \\
                              & Property Prediction                        & 709                      & 59.23                           & 40.77                           \\ \cline{2-5} 
                              & Total                                       & 4,009                     & 90.88                           & 9.12                            \\  \hline
\multirow{5}{*}{MQA}   & Preperty                                    & 6,267                     & 100.00                          & 0.00                            \\
                              & Usage                                       & 3,074                     & 100.00                          & 0.00                            \\
                              & Source                                      & 13,630                    & 100.00                          & 0.00                            \\
                              & Structure                                   & 38,603                    & 83.62                           & 16.38                           \\ \cline{2-5}
                              & Total                                       & 61,574                    & 91.19                           & 8.81                            \\ \hline
\multirow{6}{*}{\textbf{\makecell{MolGround\\(ours)}}}    & Name Entity (CNER)    &15,033                    & 0.00                            & 100.00                          \\
                              & Name2Struct (BNSM) & 1,397                      & 100.00                          & 0.00                            \\
                              & Localization (RSL) & 48,107                    & 0.00                            & 100.00                          \\
                              & Grounding (SRG)   & 13,994                    & 0.00                            & 100.00                          \\
                              & Analysis (SFA)       & 38,780                    & 0.00                            & 100.00                          \\ \cline{2-5} 
                              & Total                                       & \textbf{117,311}                    & 1.19                            & \textbf{98.81}    \\ \bottomrule                          
\end{tabular}
}
\caption{Question-answer pair distribution across descriptive (Des.) and referential (Ref.) perceptions, covering benchmarks ChemBench4K (CB4), MoleculeQA (MQA), and MolGround (Ours).}
\label{Tab:descriptiveVSreferential}
% \vspace{-0.15in}
\end{table}
% existing benchmarks in comparision to MolGround

\section{Molecular Grounding Tasks}

\begin{table*}[]
    \resizebox{1\linewidth}{!}{

\begin{tabular}{l|l|l|l}
\bottomrule
\textbf{Cognitive   levels} & \textbf{Tasks}                & \textbf{Challenges}                              & \textbf{Required Abilities}                                        \\ \hline
Remember                   & CNER                          & Diverse Entity Forms                             & Chemical Knowledge Recall; Syntax   Understanding \\ \hline
Understand                 & BNSM                          & Multimodal Transformation;                       & Semantic Understanding; Syntax Understanding     \\ \hline
Apply                      & SFA-Atom                     & Multi-instances; Coreference Resolution          & Substructure Matching; Pattern Recognition       \\ \hline
\multirow{5}{*}{Analyze}   & SFA-Heteroatoms Type & Similarity Differentiation                 & Semantic Understanding; Categorization          \\
 & SFA-Monocyclic Ring Type      & Structural Similarity;              & Pattern Recognition; Categorization  
                                      \\
                           & SFA-Non-exist Ring   & Absence Detection                  & Negative Pattern Recognition;                    \\
                           & SRG                           & Multidimensional Relations                       & Structural Understanding; Relationship Inference \\
                           & Singular RSL                  & Multi-instances & Spatial Reasoning; Pattern Recognition           \\ \hline
\multirow{3}{*}{Evaluate}  & SFA-Ring                      & Multiple Representation forms                    & Structural Comparison; Quantitative Analysis     \\
                           & SFA-Substructure              & Structural Variability;                          & Pattern Recognition; Logical Deduction           \\
                           & Multiple RSL                  & Multiple Structures; Diverse Relation Types           & Contextual Reasoning      \\ \toprule                                         
\end{tabular}
}
\caption{Grounding tasks with Bloom's Cognitive Levels, corresponding challenges, and required abilities.}
\label{TasksAndChallges}
% \vspace{-0.25in}
\end{table*}

We define 5 groups of grounding tasks by aligning to the common conventions in NLP, cheminformatics, and molecular science.
The alignment and challenges of each task are summarized in Table~\ref{TasksAndChallges}.

\begin{figure}[]  
\centering  
  \centering  
  \includegraphics[width=0.98\linewidth]{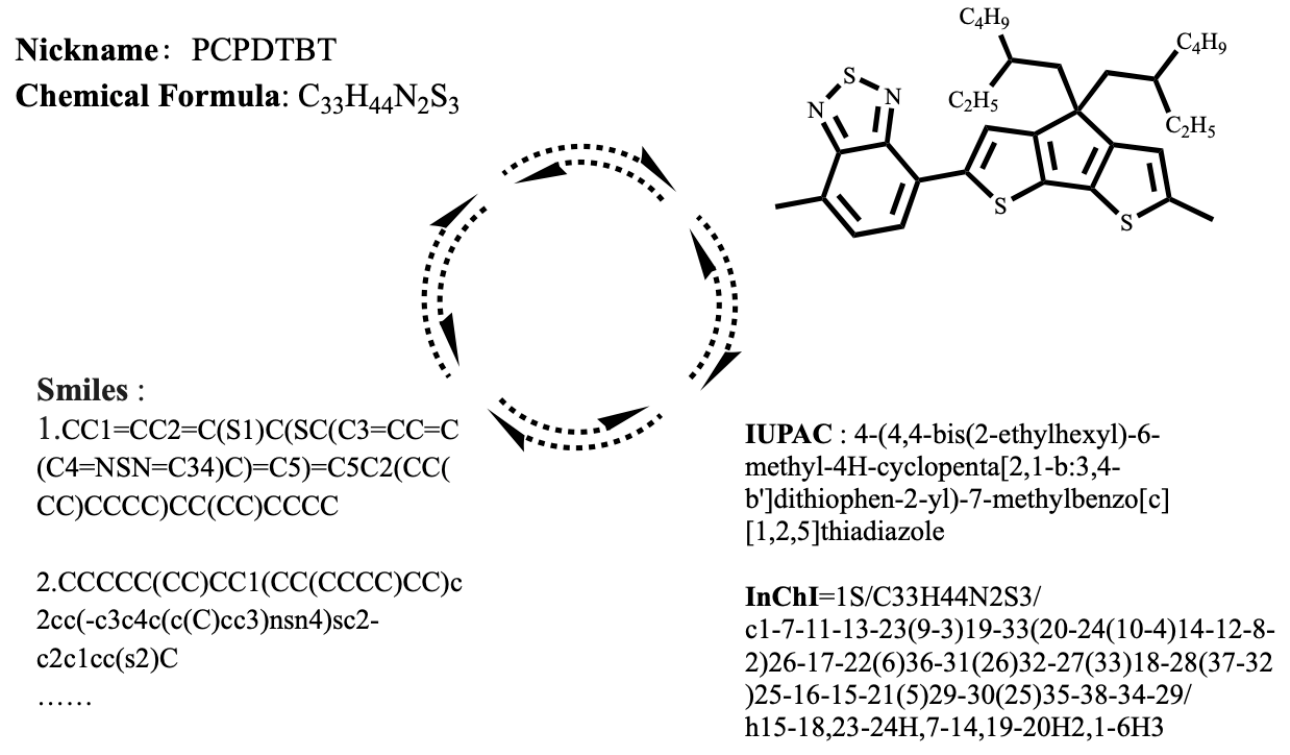}  
    \caption{Diversity in naming conventions and multimodal gaps between the textual and structural forms. }   
    \label{fig:BNSM}
    % \vspace{-0.15in}
\end{figure} 

\noindent\textbf{Chemical Named Entity Recognition (CNER)}: \textit{Recognize and extract chemical entity names (e.g., molecule names, substructure names, or functional groups) as a set $\mathcal{N}$ from a given caption $\mathcal{X}$ as}
\begin{align}
f_N:\,\,&\mathcal{X}\mapsto\,\mathcal{N}\nonumber\\
    &\mathcal{X}=\{x_i\},\,\mathcal{N}=\{n_j\}
\end{align}
where $x_i$ is the $i^{th}$ token in the sequence $\mathcal{X}$, and $n_j$ is a quadruple $(c_j,b_j,l_j,r_j)$ consisting of the $j^{th}$ extracted name entity $c_j$ and its role $r_j\in\mathcal{R}$, beginning position $0\leq b_j \leq \|X\|$, and length $l_j$. 
Note $\mathcal{R}$ is a set of predefined roles (e.g., donor, acceptor) which are contextual and application-specific.

% Challenge: Diversity of Naming Convention , Context sensitivity 
This task reflects referential perception by linking textual mentions of chemical entities to their semantic roles and serves as a foundation for molecular grounding by identifying key entities for downstream tasks.
While similar to Named Entity Recognition (NER) in NLP, CNER extends the task by additionally identifying the roles of extracted entities.
Furthermore, unlike NER, where entities are typically proper nouns or noun phrases, chemical entities are significantly more diverse and technically complex.
For instance, the \textit{drug acetaminophen} exemplifies this complexity: it has multiple IUPAC names, such as \textit{N-(4-hydroxyphenyl)acetamide}, \textit{4'-hydroxyacetanilide}, and \textit{p-hydroxyacetanilide}; a molecular formula, \textit{C8H9NO2}; an InChI representation, \textit{InChI=1S/C8H9NO2/c1-6(10)9-7-2-4-8(11)5-3-7/h2-5,11H,1H3,(H,9,10)}; as well as various SMILES representations and trade names like \textit{Tylenol}, \textit{Panadol}, and \textit{Calpol}.
As illustrated in Figure~\ref{fig:BNSM}, a CNER model must accommodate these diverse forms, demanding the ability to recall chemical domain knowledge and a deep understanding of chemical syntax and representation conventions.

\noindent\textbf{Bidirectional Name-Structure Mapping (BNSM)}: \textit{Translate chemical names $\mathcal{N}$ into corresponding structural representations (e.g., SMILES, InChI, molecular graphs) $\mathcal{S}$ or convert given structural representations back into their corresponding names as}
\begin{align}
f_{n2s}:\,\,&\mathcal{N}\mapsto\,\mathcal{S}\nonumber\\
f_{s2n}:\,\,    &\mathcal{S}\mapsto\,\mathcal{N}\nonumber\\
    &\mathcal{N}=\{n_i\},\,\mathcal{S}=\{s_j\}
\end{align}
where structural representation $\mathcal{S}$ is sequences of textual codes in SMILES, InChI, or molecular graphs wrapping atoms (nodes) and bonds (edges).

This task bridges textual and structural representations, embodying referential perception by grounding a molecule's name to its physical structure and vice versa. 
It aligns with translation tasks in NLP and structure-based prediction tasks in cheminformatics.
%
% challenge: multimodal, error tolerance
%
As illustrated in Figure~\ref{fig:BNSM}, unlike the sequence-to-sequence framework used in NLP translation, this task introduces an additional multimodal challenge. 
This complexity arises from the hierarchical and graph-based nature of molecular structures, which are governed by spatial and chemical constraints. 
Furthermore, this task has extremely low error tolerance, as even a minor mistake in structure representation can lead to a fundamentally different molecule (e.g., \textit{C1=CC=CC=C1} vs. \textit{C1=NC=CC=C1}).

\begin{figure}[]  
\centering  
  \centering  
  \includegraphics[width=0.65\linewidth]{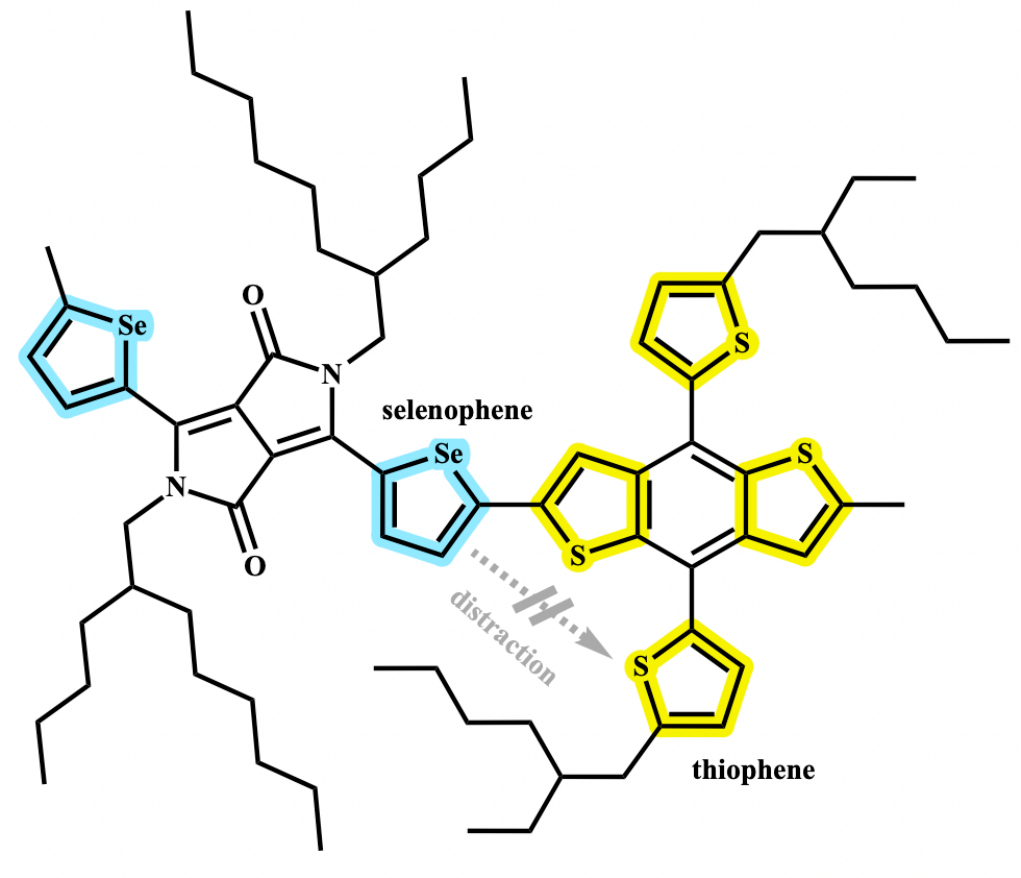}  
    \caption{Multiple instances of \textit{thiophene} rings, varying by rotations, present a challenge in identifying a generalizable feature for localization. Additionally, \textit{selenophene} rings, differing by only one atom from \textit{thiophene}, may further complicate localization.}   
    \label{fig:RSL}
    % \vspace{-0.2in}
\end{figure} 

\noindent\textbf{Referential Substructure Localization (RSL)}: \textit{Identify the specific occurrences of substructures (e.g., functional groups, rings, or atoms) within a molecule's structural representation $\mathcal{G}$, based on their names or descriptions $\mathcal{N}$ as}
\begin{align}
f_{L}:\,\,&(\mathcal{N},\mathcal{G})\mapsto\,\mathcal{L}\nonumber\\
    &\mathcal{N}=\{n_i\},\,\mathcal{G}=(\mathcal{V},\mathcal{E}),\,\mathcal{G}_i\subseteq\mathcal{G},\nonumber\\
    &\mathcal{L}=\{\mathcal{L}_i\}\in \{\mathcal{G}_i\}\times\mathcal{G}
\end{align}
where $\mathcal{V}$ is the set of atoms (nodes) and
$\mathcal{E}$ is the set of bonds (edges), $\mathcal{G}_i=(\mathcal{V}_i,\mathcal{E}_i)$ is the substructure graph for $n_i$, and $\mathcal{L}_i$ is the location indicator for $\mathcal{G}_i$ within the molecular graph $\mathcal{G}$.
$\mathcal{L}_i=(\mathcal{L}^{atom}_i,\mathcal{L}^{bond}_i)$ consists of indices of $\mathcal{G}_i$'s atoms and bonds within the molecular graph $\mathcal{G}$, where $\mathcal{L}^{atom}_i=\{m\vert v_m\in \mathcal{V}_i\}$ and $\mathcal{L}^{bond}_i=\{(m,n)\vert (v_m,v_n)\in\mathcal{E}_i\}$.

This task emphasizes referential perception by mapping textual or conceptual references to their precise structural counterparts. 
It is analogous to object detection in vision and token-level alignment in NLP.
%
% multiple instances, and multiple distractors
Building upon CNER and BNSM, the new challenge imposed in RSL is the existence of multiple instances of the target and possible distractors.
Those distractors are often with similar structures as the target, further challenging the low tolerance at fine grained level.
Examples can be found in Figure~\ref{fig:RSL}.

\noindent\textbf{Substructure Relationship Grounding (SRG)}: \textit{Identify the relationships (e.g., composition, directed attachment, functional integration, or coexistence) between substructures within a molecule, as represented by a caption $\mathcal{X}$ and the corresponding molecular graph $\mathcal{G}$ as}
\begin{align}
f_{K}:\,\,&(\mathcal{X},\mathcal{G})\mapsto\,\mathcal{K}\nonumber\\
    &\mathcal{X}=\{x_i\},\,\mathcal{G}=(\mathcal{V},\mathcal{E}),\,\mathcal{G}_i,\mathcal{G}_j\subseteq\mathcal{G},\nonumber\\&\mathcal{K}=\{k_{ij}\}\in \{\mathcal{G}_i\}\times\{\mathcal{G}_j\}
\end{align}
where $\mathcal{G}_i$ and $\mathcal{G}_j$ are the $i^{th}$ and $j^{th}$ substructure graphs, and $k_{ij}$ is their identified relationship.

\begin{figure}[]  
\centering  
  \centering  
  \includegraphics[width=0.7\linewidth]{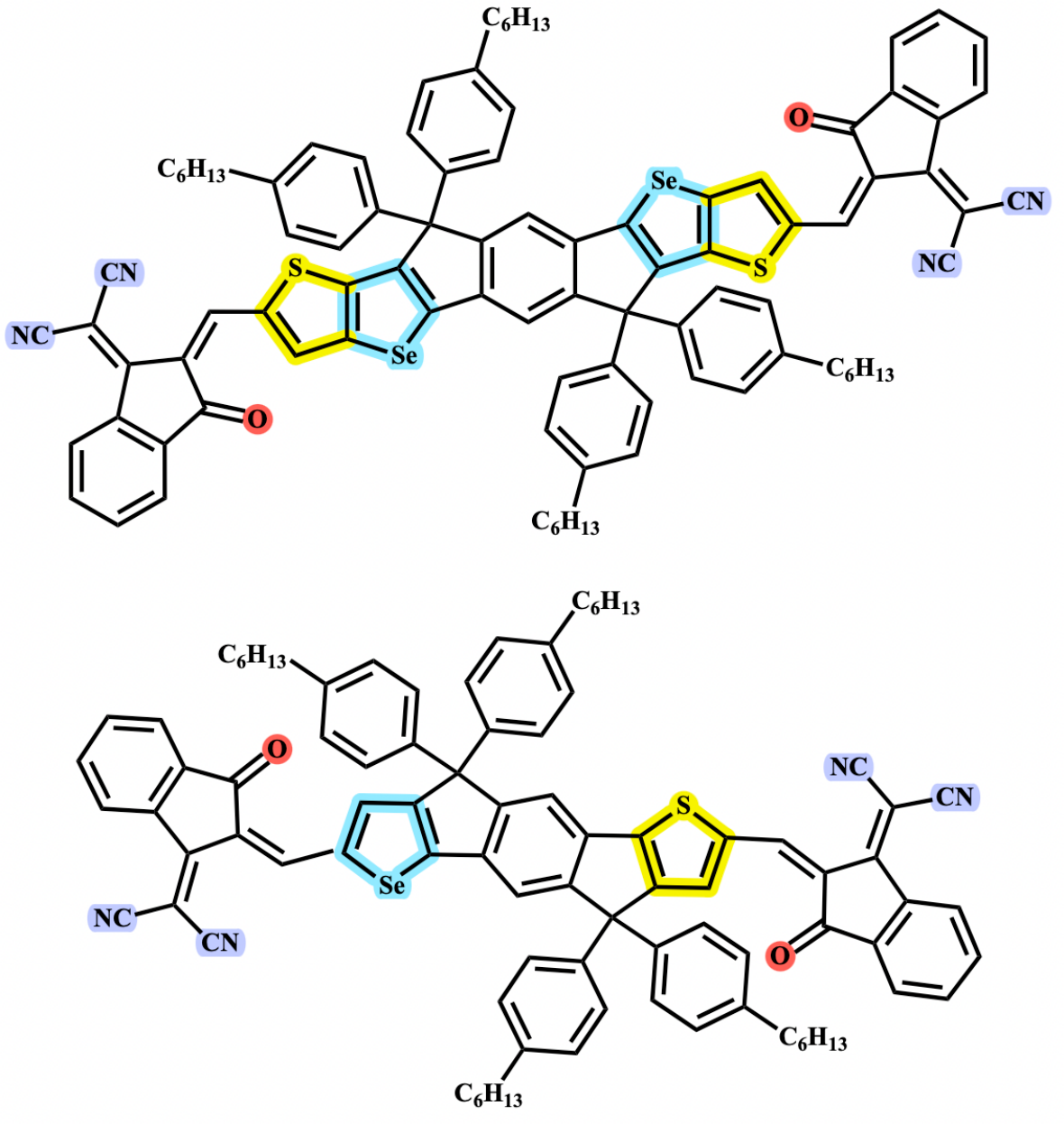}  
    \caption{The relationship of ``functional integration'' between the \textit{thiophene} (yellow) and \textit{selenophene} (blue) rings varies significantly across different molecules (attached in one case but distinctly separate in another).}   
    \label{fig:SRG}
    % \vspace{-0.2in}
\end{figure}

This task builds on referential perception by modeling the interactions and dependencies between molecular substructures, providing insights into their functional roles. 
It draws parallels to relation extraction in NLP and interaction modeling in molecular sciences.
%
% challenge: multidimensional, intricately intertwined with their associated physical factors (need a term)
The key challenge in this task lies in the multidimensional nature of the relationships. 
Unlike conventional NLP relation extraction, which is primarily governed by semantic correlations between entities, SRG relationships are multidimensional, incorporating chemical, spatial, physical, and hierarchical factors.
More specifically, this complexity means that chemical relationships, such as composition, directed attachment, or functional integration, are intricately intertwined with their associated physical factors.
This contrasts sharply with NLP relations (e.g., is-a, is-part-of), which are often straightforwardly defined.
Figure~\ref{fig:SRG} illustrates this challenge.

\noindent\textbf{Substructure Frequency Analysis (SFA)}: \textit{Count the number of occurrences of a specified substructure (indicated by its name $n_i\in\mathcal{N}$) within the structural representation $\mathcal{G}$ of a given molecule as}
\begin{align}
f_{F}:\,\,&(\mathcal{N},\mathcal{G})\mapsto\,\mathcal{F}\nonumber\\
    &\mathcal{N}=\{N_i\},\,\mathcal{G}=(\mathcal{V},\mathcal{E}),\,\mathcal{G}_i\subseteq\mathcal{G},\nonumber\\&\mathcal{F}=\{k_{i}\}\in \mathbb{N}
\end{align}
where $k_i$ is the frequency of $n_i$ counted by the occurrences of its substructure graph $\mathcal{
G
}_i$ within $\mathcal{G}$.

This task extends referential perception by quantifying the presence of referenced substructures, supporting downstream molecular grounding tasks such as property prediction or functional analysis. 
It aligns with token frequency counting in NLP and motif detection in cheminformatics.
However, this goes beyond a simple counting task. 
The complexity arising from multiple representation forms, hierarchical definitions, multidimensionality, and multiresolution makes the target of counting dynamic and context-dependent, unlike the fixed nature of token frequency analysis in NLP.

\section{Benchmarking}
Benchmarking in the chemical domain is expensive, largely due to its heavy reliance on human expertise.  
To build the largest molecular understanding benchmark to date, we adopt an interactive approach based on the Spiral Model \cite{BoehmSpiralModel}.  
Specifically, we develop a prototype of a grounding agent to facilitate the process.  
The agent automates data collection, cleaning, and structuring, after which the data is validated, corrected, or filtered by human experts.  
Data entries rejected by human experts are refined by the agent and resubmitted for further review.  
This iterative interaction between humans and the agent continues until convergence is achieved.  
Throughout this process, the agent itself is iteratively improved as part of an exploration into effective grounding methodologies, while simultaneously enhancing both the scale and quality of the benchmark.

\subsection{Grounding Agent Prototyping}
Our study shows that the most effective approach is a multi-agent system composed of a meta-retriever, an LLM-based text interpreter, and a structure parser.  
The meta-retriever is built using PubChem APIs \cite{PubChem} and is responsible for collecting molecular names, properties, and descriptions. 
The text interpreter leverages large language models (LLMs) to perform named entity recognition and relationship analysis at the textual level.
The structure parser is developed using RDKit\footnote{RDKit: Open-source cheminformatics, https://www.rdkit.org} and handles structure retrieval, comparison, and validation.  

These three agents work collaboratively: the meta-retriever gathers metadata as needed and provides it to the text interpreter as examples or contextual information for in-context learning.
The text interpreter extracts names and relationships from captions, passing them to the structure parser, which converts the information into molecular structures. 
The structure parser then compares or validates the structures to produce grounded outputs.  
This process effectively handles all five grounding tasks. 
%
% Due to space constraints, further details are available in the source code at \textcolor{magenta}{\url{https://github.com/open_upon_acceptance}}.

% As the existing models struggle in referential substructure localization, we explore a possible solution for molecular grounding tasks by proposing a grounding agent. The grounding agent uses a linguistic tool for semantic understanding and semantic-to-structure transformation as well as a sub-graph retrieval tool for substructure locations.

% The workflow of the ground agent stimulates the heuristic steps outlined in Figure \ref{fig:groundingProcess}. 
% Specifically, the agent process the input through the following steps and output a executable code: 
% 1. Chemical named entity recognition and relations analysis
% 2. Chemical names to SMILES strings transformation
% 3. Molecule and substructure SMILES parsing using RDKIT
% 4. Substructure matching using RDKIT
% 5. Generate the result following the given format.
% 6. Output python code
% The agent is asked to generate a python code containing the output of step 2 as a variable and execute step 3-5 to generate the grounding outputs.
% The python code is executed to get the result.
% The detailed prompt to defined the grounding agent is provided in the supplementary material. 

\subsection{Data Collection and Preprocessing}
We collected molecules from existing molecular captioning datasets, such as ChEBI-20 \cite{ChEBI-20text2mol} and LPM-24 \cite{ACLMoleculeCaptionWorkshop}.
Additionally, we extended our collection with molecules published in chemical literature \cite{OPV_molecule_source}.
In total, this resulted in a dataset of 55,989 molecules.
The collected molecules exhibit varying levels of structural complexity. 
Specifically, the number of atoms per molecule ranges from 1 to 574, with a median value of 33. 
The number of rings varies from 0 to 69, while the number of bonds spans from 1 to 642.

For molecules lacking captions (approximately 2\% of the dataset), we utilized GPT-4o \cite{GPT4o2024openai} to generate detailed captions.
This was achieved by inputting the molecule's IUPAC name, SMILES representation, relevant literature, and molecular structure image into GPT-4o, along with prompt templates designed by chemical experts (details provided in the Appendix \ref{Appendix_prompt}).
The templates were tailored to generate substructure-focused content, such as identifying the substructures within a molecule, describing how they are connected, and outlining their properties.
As a result, we constructed a dataset of 55,989 molecule-caption pairs.

\subsection{Structurization and Annotation}
The structuring and annotation process is performed iteratively, allowing our grounding agent to collaborate with human chemical experts.
The 117k QA pairs listed in Table \ref{Tab:descriptiveVSreferential} are constructed based on the human annotation data. We split the QAs into training, validation, and testing sets using a split ratio (80\%, 10\%, 10\%) on each task, and ensure that there is no overlapped molecule between different sets.

\section{Experiments}

% Please add the following required packages to your document preamble:
% \usepackage{multirow}
\begin{table*}[]
\small
\small{
\newcolumntype{C}[1]{>{\centering\arraybackslash}p{#1}}
    % \resizebox{1\textwidth}{!}{
% \begin{tabular}{C{2.5cm}|C{2.6cm}|C{1.2cm}|C{1.2cm}|C{1.2cm}|C{1.2cm}|C{1.2cm}|C{1.2cm}}
\begin{tabular}{C{0.9cm}|C{2.2cm}|C{0.8cm}|C{0.8cm}|C{0.6cm}|C{0.6cm}|C{0.1m}C{0.3cm}C{0.6cm}|C{0.1cm}C{0.3cm}C{0.3cm}C{0.5cm}}
\bottomrule
                               & \textbf{Tasks}            & \textbf{CNER}           & \textbf{BNSM}           & \textbf{SRG}            & \textbf{SFA}            & \multicolumn{3}{c|}{\textbf{S-RSL}}                                                          & \multicolumn{4}{c}{\textbf{M-RSL}}                                                                          \\ \cline{2-13}
                               & Metric           & F1             & Acc.       & F1        & Acc.        & \multicolumn{1}{c|}{$F1_l$}         & \multicolumn{1}{c|}{$IoU_l$}        & $Acc_g$        & \multicolumn{1}{c|}{$F1_l$}         & \multicolumn{1}{c|}{$IoU_l$} & \multicolumn{1}{c|}{$Acc_g$} & $Cov_s$  \\ \hline
\multirow{7}{*}{\textbf{LLM}}  & GPT-4o            & \textbf{0.633}          & 0.125 &  \textbf{0.160} & \textbf{0.337} & \multicolumn{1}{c|}{0.015}          & \multicolumn{1}{c|}{0.148}          & \textbf{0.755} & \multicolumn{1}{c|}{\textbf{0.011}} & \multicolumn{1}{c|}{\textbf{0.073}}   & \multicolumn{1}{c|}{\textbf{0.429}}   & \textbf{0.543} \\
                               & LLaMA  3.1-8B      & 0.504          & 0.092          &  0.001        & 0.111          & \multicolumn{1}{c|}{0.006}          & \multicolumn{1}{c|}{\textbf{0.175}} & 0.672          & \multicolumn{1}{c|}{0.000}          & \multicolumn{1}{c|}{0.018}   & \multicolumn{1}{c|}{0.083}   & 0.126 \\
                               & LLaMA  3.1-70B     &   0.637      & 0.063          &    0.004      & 0.233          & \multicolumn{1}{c|}{0.008}          & \multicolumn{1}{c|}{0.100}          & 0.473          & \multicolumn{1}{c|}{0.001}          & \multicolumn{1}{c|}{0.009}   & \multicolumn{1}{c|}{0.031}   & 0.045 \\
                               & BioT5+           & 0.000          & 0.000          &  0.000         & 0.000          & \multicolumn{1}{c|}{0.000}          & \multicolumn{1}{c|}{0.000}          & 0.000          & \multicolumn{1}{c|}{0.000}          & \multicolumn{1}{c|}{0.000}   & \multicolumn{1}{c|}{0.000}   & 0.000    \\
                               & ChemLLM-7B       & 0.000         & 0.000          &  0.000         & 0.154          & \multicolumn{1}{c|}{0.000}          & \multicolumn{1}{c|}{0.000}          & 0.678          & \multicolumn{1}{c|}{0.000}          & \multicolumn{1}{c|}{0.000}   & \multicolumn{1}{c|}{0.000}   & 0.000    \\
                               & Mol-Instructions & 0.152          & 0.000          &  0.000         & 0.000          & \multicolumn{1}{c|}{0.000}          & \multicolumn{1}{c|}{0.000}          & 0.000          & \multicolumn{1}{c|}{0.000}          & \multicolumn{1}{c|}{0.000}   & \multicolumn{1}{c|}{0.000}   & 0.000    \\ 
                               
% &nach0 \cite{Micha2024nach0} & \\
                               
                               \hline
\multirow{2}{*}{\textbf{MLLM}} & GPT-4o-Vision     & 0.578  & \textbf{0.246}         & 0.061 & 0.321          & \multicolumn{1}{c|}{0.004}          & \multicolumn{1}{c|}{0.052}          & 0.332          & \multicolumn{1}{c|}{0.000}          & \multicolumn{1}{c|}{0.001}   & \multicolumn{1}{c|}{0.012}   & 0.016    \\ 
                               & LLaVA-Next-7B    &   0.412   & 0.021          &     0.042       & 0.142          & \multicolumn{1}{c|}{\textbf{0.020}} & \multicolumn{1}{c|}{0.174}          & 0.737          & \multicolumn{1}{c|}{0.000}          & \multicolumn{1}{c|}{0.001}   & \multicolumn{1}{c|}{0.004}   & 0.005   \\ \toprule
\end{tabular}
}
% }
\caption{Comparison of LLMs and MLLMs performance across five molecular grounding tasks. For RSL, results are reported at both singular (S-RSL) and multiple (M-RSL) substructure levels.}
\label{tab:perfOn4Levels}
% \vspace{-0.2in}
\end{table*}

\subsection{Baselines}
We employ 8 LLMs as baselines, including general-domain models like GPT4o \cite{GPT4o2024openai} and LLaMA 3.1 (8B and 70B) \cite{grattafiori2024llama3herdmodels}, as well as models specifically tailored for molecular understanding, such as Bio-T5+ \cite{bioT52024ACL}, ChemLLM (7B) \cite{Zhang2024ChemLLMAC}, and Mol-Instructions \cite{fang2024molinstructions}.  
Furthermore, we investigate LLM learning techniques, including In-Context Learning (ICL), such as Retrieval-Augmented Generation (RAG), and Supervised Fine-tuning (SFT) using LoRA \cite{hu2022lora}.  
Given that molecular structures can be represented as graphs, we also incorporate Multi-modal LLMs (MLLMs) like GPT4o Vision \cite{GPT4o2024openai} and LLaVA-Next \cite{liu2023llava} in our evaluations.

\begin{table*}[t]
\small
\centering
\newcolumntype{C}[1]{>{\centering\arraybackslash}p{#1}}
    % \resizebox{1\textwidth}{!}{
\begin{tabular}{p{4cm}|C{3.0cm}|C{0.9cm}|C{0.9cm}|C{0.9cm}|C{0.9cm}|C{0.9cm}|C{1.2cm}}
\bottomrule
% Please add the following required packages to your document preamble:
% \usepackage{multirow}

% \bottomrule
% Please add the following required packages to your document preamble:
% \usepackage{multirow}

\multicolumn{1}{c|}{}                    & \textbf{Tasks}            & \textbf{CNER}           & \textbf{BNSM}           & \textbf{SRG}            & \textbf{SFA}            & \textbf{S-RSL}   & \textbf{M-RSL}   \\ \cline{2-8}
\multicolumn{1}{c|}{}                    & Metric           & F1             & Acc.      & F1         & Acc.         & $F1_l$         & $F1_l$         \\ \hline
\multirow{3}{*}{\textbf{Baselines}}     & GPT-4o            & 0.633         & 0.125          & 0.160 & 0.337          & 0.015          & 0.011          \\
                                          & LLaMA  3.1-8B      & 0.504          & 0.092          & 0.001         & 0.111          & 0.006          & 0.000          \\
                                          & Mol-Instructions & 0.152          & 0.000          &     0.000     & 0.001          & 0.000          & 0.000          \\ \hline
\multirow{3}{*}{\textbf{Baselines + ICL (Few-shot)}} & GPT-4o            & 0.721          &  0.314&     \textbf{0.385}       & 0.754          & 0.017          & 0.270          \\
                                          & LLaMA  3.1-8B      & 0.722         & 0.180          &    0.195       & 0.587          & 0.003          & 0.120          \\
                                          & Mol-Instructions & 0.350         & 0.000          &  0.000       & 0.117          & 0.000          & 0.036          \\ \hline
\multirow{3}{*}{\textbf{Baselines + ICL (RAG)}}      & GPT-4o            &   \textbf{0.915}        & 0.397 &  0.368     & 0.754          & 0.174          & 0.171          \\
                                          & LLaMA  3.1-8B      & 0.881           & 0.361          &  0.072     & 0.655          & 0.149          & 0.113          \\
                                          & Mol-Instructions & 0.864           & 0.083          &    0.000      & 0.561          & 0.091          & 0.072          \\ \hline
\multirow{2}{*}{\textbf{Baselines + SFT}}           & LLaMA  3.1-8B      & 0.641  & \textbf{0.426}          &   0.078       & 0.899          & 0.275          & 0.315          \\
                                          & Mol-Instructions &   0.727         & 0.397          &    0.072       & \textbf{0.917} & \textbf{0.295}	&\textbf{0.337} \\
\toprule
\end{tabular}
% }
\caption{Performance improvement through the integration of ICL and SFT techniques across five grounding tasks. For RSL, results are presented at both singular (S-RSL) and multiple (M-RSL) substructure levels.}
\label{tab:perfOnICLandSFT}
% \vspace{-0.1in}
\end{table*}

\subsection{Evaluation of Pretrained Models}

\begin{table*}[t]
\small
\centering
% \renewcommand{\arraystretch}{1.0}
%     \resizebox{1\textwidth}{!}{
\newcolumntype{C}[1]{>{\centering\arraybackslash}p{#1}}

% \begin{tabular}{C{2cm}|C{0.9cm}|C{0.9cm}|C{0.9cm}|C{0.9cm}}
\begin{tabular}{C{2.3cm}|C{3.2cm}|C{0.6cm}|C{0.6cm}|C{0.6cm}|C{0.7cm}|C{0.6cm}|C{0.6cm}|C{0.6cm}|C{0.6cm}|C{0.7cm}}

\bottomrule
\multicolumn{1}{c|}{}                      & \multicolumn{1}{c|}{\textbf{Tasks}}         & \multicolumn{4}{c|}{\textbf{S-RSL}}                                  & \multicolumn{5}{c}{\textbf{M-RSL}}                                                   \\ \cline{2-11}
\textbf{}                                 & Metic                             & $F1_l$         & $IoU_l$        & $IoU_g$        & $Acc_g$        & $F1_l$         & $IoU_l$        & $IoU_g$        & $Acc_g$        & $Cov_s$        \\ \hline
\multirow{3}{*}{\textbf{Pre-trained}}     & GPT-4o                             & 0.015          & 0.148          & 0.135          & 0.755          & 0.011          & 0.073          & 0.054          & 0.429          & 0.543          \\
                                          & LLaMA  3.1-8B                       & 0.006          & 0.175          & 0.141          & 0.672          & 0.000          & 0.018          & 0.018          & 0.083          & 0.126          \\
                                          % & LLaMA  3.1-70B                      & 0.008          & 0.100          & 0.083          & 0.473          & 0.001          & 0.009          & 0.006          & 0.031          & 0.045          \\
                                          & Mol-Instruct-8B                   & 0.000          & 0.000          & 0.000          & 0.000          & 0.000          & 0.000          & 0.000          & 0.000          & 0.000          \\ \hline
\multirow{4}{*}{\textbf{ICL(Few-shot)}}   & GPT4o                             & 0.017          & 0.059          & 0.143          & 0.291          & 0.270          & 0.454          & 0.399          & \textbf{0.779} & \textbf{0.905} \\
                                          & LLaMA  3.1-8B                       & 0.003          & 0.014          & 0.052          & 0.102          & 0.120          & 0.207          & 0.181          & 0.379          & 0.432          \\
                                          % & LLaMA  3.1-70B                      & 0.000          & 0.029          & 0.114          & 0.234          & 0.228          & 0.381          & 0.330          & 0.652          & 0.758          \\
                                          & Mol-Instruct-8B                   & 0.000          & 0.000          & 0.000          & 0.000          & 0.036          & 0.079          & 0.049          & 0.117          & 0.143          \\ \hline
\multirow{4}{*}{\textbf{ICL(RAG)}}        & GPT4o                             & 0.174          & 0.361          & 0.337          & 0.415          & 0.171          & 0.301          & 0.290          & 0.369          & 0.718          \\
                                          & LLaMA  3.1-8B                       & 0.149          & 0.332          & 0.307          & 0.387          & 0.113          & 0.262          & 0.255          & 0.344          & 0.760          \\
                                          % & LLaMA  3.1-70B                      & 0.099          & 0.251          & 0.233          & 0.301          & 0.109          & 0.229          & 0.225          & 0.303          & 0.702          \\
                                          & Mol-Instruct-8B                   & 0.091          & 0.284          & 0.243          & 0.326          & 0.072          & 0.203          & 0.183          & 0.256          & 0.637          \\ \hline
\multirow{3}{*}{\textbf{SFT}}             & LLaMA  3.1-8B                       & 0.275          & 0.483          & 0.466          & 0.548          & 0.315          & 0.515          & 0.487          & 0.587          & 0.850          \\
                                          % & LLaMA  3.1-70B                      & 0.010          & 0.113          & 0.091          & 0.143          & 0.000          & 0.007          & 0.010          & 0.015          & 0.024          \\
                                          & Mol-Instruct-8B                   & 0.295          & 0.499          & 0.486          & 0.565          & 0.337          & 0.518          & 0.493          & 0.586          & 0.833          \\ \hline
\multirow{3}{*}{\textbf{MLLM}}            & GPT4o                             & 0.004          & 0.052          & 0.054          & 0.332          & 0.000          & 0.001          & 0.001          & 0.012          & 0.016          \\
                                          & LLaVA-Next-7B & 0.020          & 0.174          & 0.112          & 0.737          & 0.000          & 0.001          & 0.001          & 0.004          & 0.005          \\
                                          & LLaMA  3.2-11B-Vision               & 0.010          & 0.113          & 0.085          & 0.555          & 0.003          & 0.062          & 0.046          & 0.280          & 0.402          \\ \hline
\multirow{4}{*}{\textbf{Grounding Agent}} & GPT4o                             & \textbf{0.630} & \textbf{0.685} & \textbf{0.647} & \textbf{0.933} & \textbf{0.541} & \textbf{0.566} & \textbf{0.546} & 0.776          & 0.818          \\
                                          & LLaMA  3.1-8B                       & 0.334          & 0.383          & 0.364          & 0.863          & 0.426          & 0.527          & 0.448          & 0.580          & 0.688          \\
                                          % & LLaMA  3.1-70B                      & 0.503          & 0.545          & 0.514          & 0.921          & 0.500          & 0.529          & 0.507          & \textbf{0.779} & 0.834          \\
                                          & Mol-Instruct-8B                   & 0.000          & 0.000          & 0.000          & 0.000          & 0.311          & 0.446          & 0.310          & 0.356          & 0.382  \\ \toprule       
\end{tabular}
% }
\caption{Performance of grounding agents with different backbones. Both local ($_l$) and global ($_g$) RSL are reported.}
\label{tab:perfOnGrounding}
% \vspace{-0.2in}
\end{table*}

Table \ref{tab:perfOn4Levels} compares the performance of LLMs and MLLMs across five molecular grounding tasks.
Overall, most tasks remain challenging for all baseline models, with accuracy generally below 0.5.
In particular, BNSM, SFA, and RSL prove to be the most difficult, with all models achieving accuracies below 0.337.
By contrast, CNER and SRG exhibit relatively better performance. The highest F1-score for CNER is 0.633 and the highest accuracy for SRG  is 0.803, which are achieved by GPT-4o.
For the CNER task, models often over-extract functionality words as chemical substructure names (e.g., ``amide functionality'') and struggle to correctly assign roles to extracted chemical names.
% leading to low F1-score.
For the SFA, models perform well on counting non-existent rings but struggle with monocyclic ring identification due to subtle structural differences.
For example, the predicted probabilities of the benzene to be a donor or acceptor is Moreover, BioT5+ fails to follow instructions, and Mol-Instructions exhibits a strong bias in selecting question choices .
While chemical name recognition (CNER) is relatively well-handled, with most models achieving an F1-score above 0.9, smaller models (e.g., LLaMA  3.1-8B, ChemLLM-7B)  similar names (e.g., \textit{ether} vs. \textit{ester})
For more complex tasks (BNSM and SRG), performance remains low, even for the best model, GPT-4o, which achieves only 0.125 and 0.333 accuracy, respectively. In BNSM, models perform well for simple structures but struggle with complex mappings, often generating incorrect yet structurally similar names or SMILES strings. SRG results indicate difficulty in establishing substructures' relationships, with models misinterpreting textual cues and neglecting structural connections. 
RSL task witnessed a particularly poor performance, with the best F1 (0.020) and IoU (0.174). 
In singular RSL, GPT-4o achieves the highest accuracy (0.755) but with low IoU (0.148), while LLaMA 3.1-8B prioritizes broader highlighting at the cost of accuracy. For multiple RSL, most models generate invalid outputs and show low substructure coverage (max 54.3\%).
MLLMs, despite access to structural images, do not surpass LLMs (except BNSM).

\subsection{Evaluation of ICL and SFT}

Table \ref{tab:perfOnICLandSFT} examines how ICL and SFT impact the tasks. 
Overall, while both ICL and SFT improve results, the gains in RSL tasks remain limited. ICL significantly enhances tasks requiring chemical knowledge recall, particularly CNER and SFA. 
Few-shot learning boosts GPT-4o and LLaMA  3.1-8B accuracy to 99.4\% and 96.0\% on CNER, respectively. RAG dramatically improves SFA performance, with increases of 124\% for GPT-4o and 489\% for LLaMA  3.1-8B. Mol-Instructions sees an extreme gain of 56009\%, though this may indicate prior poor performance rather than true capability improvement.
SFT provides the most substantial boost for grounding tasks. For instance, LLaMA  3.1-8B's F1-score in singular RSL gains from 0.006 to 0.275, and in multiple RSL from 0.001 to 0.315. 
Mol-Instructions follows a similar trend, demonstrating SFT’s effectiveness in refining molecular representations.
However, improvements are not consistent across models. ICL techniques sometimes degrade performance. For example, Few-shot learning and RAG lower GPT-4o's accuracy on SRG, as additional examples introduce substructure relationships that distract the model.

\subsection{Evaluation of the Agent Prototype}

\begin{figure}[h]  
\renewcommand{\arraystretch}{1.0}

\centering  
  \centering  
  \includegraphics[width=1\linewidth]{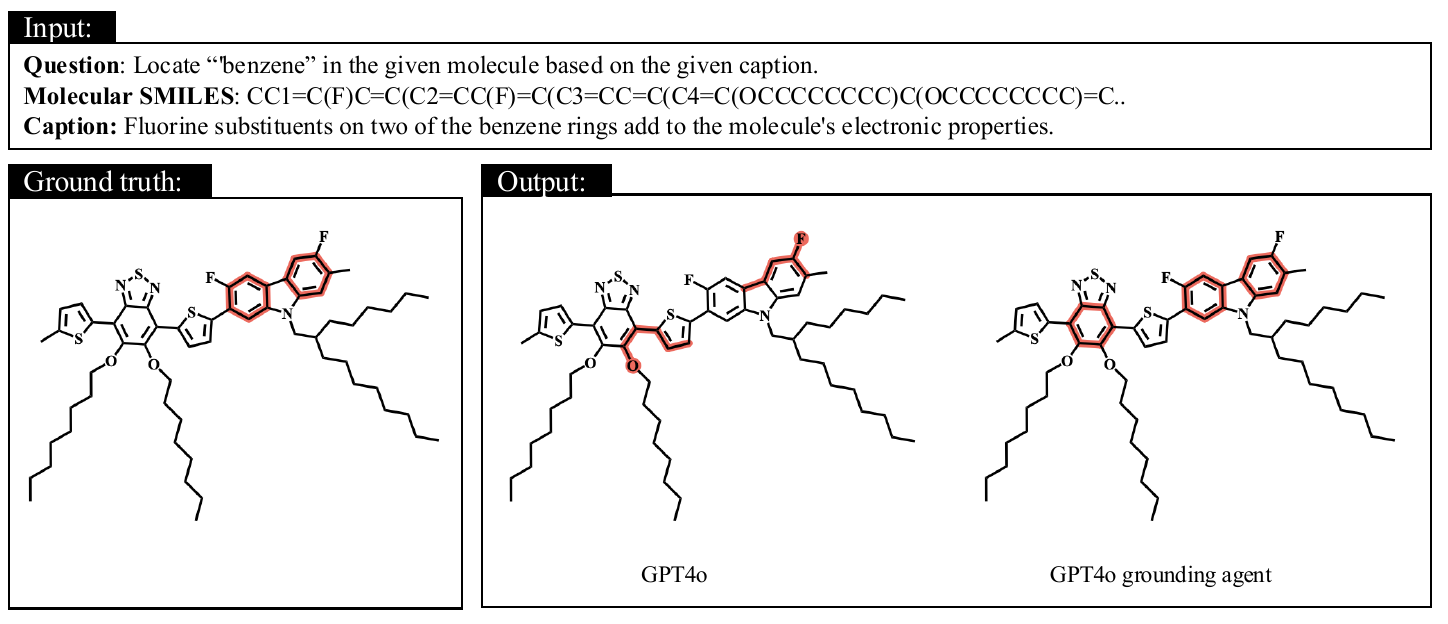}  
    \caption{Comparison of the ground truth and grounding outputs by GPT4o and the grounding agent.}   
    \label{fig:comparedAgent}
    % \vspace{-0.2in}
\end{figure}

\begin{figure}[h]  
\renewcommand{\arraystretch}{1.0}

\centering  
  \centering  
  \includegraphics[width=1\linewidth]{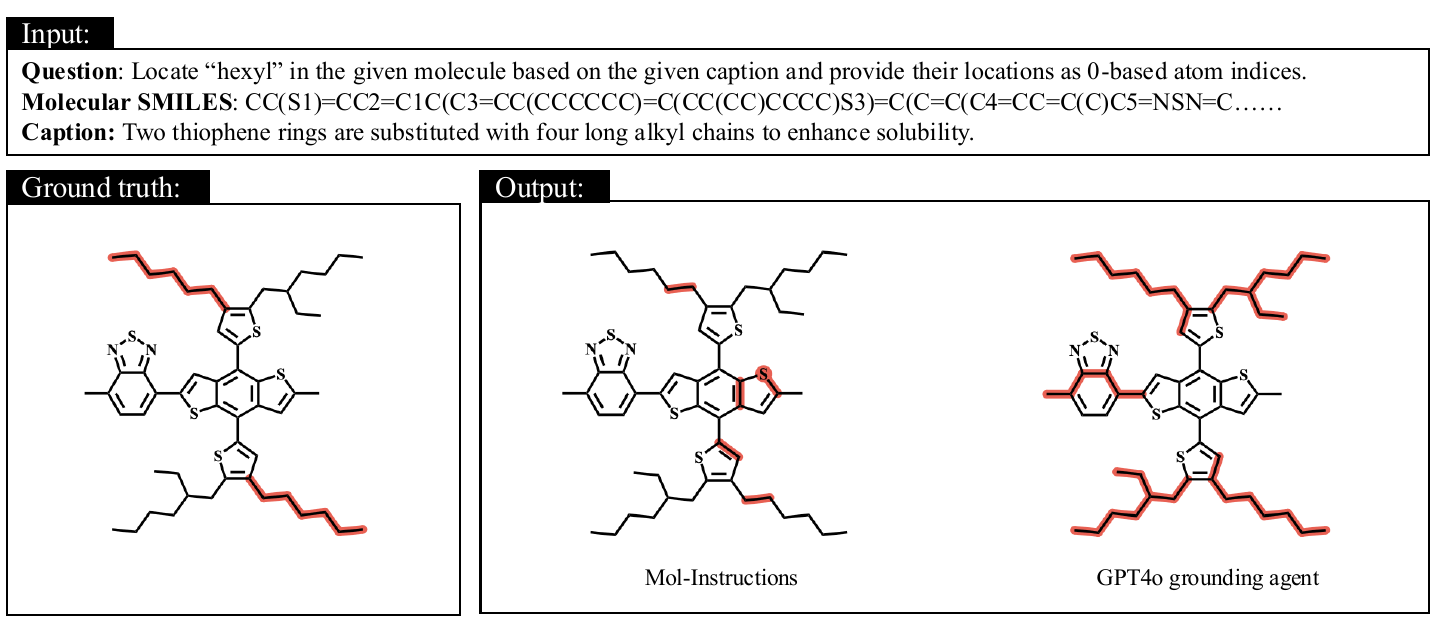}  
    \caption{Irrelevant and wrong grounding results generated by Mol-Instructions and the grounding agent.}   
    \label{fig:IrrelevantAndWrongResult}
    % \vspace{-0.2in}
\end{figure}

Table \ref{tab:perfOnGrounding} compares RSL performance of the agents and other models. The grounding agents outperform other models in singular and multiple RSL tasks across most metrics, except for substructure coverage. Their advantage primarily stems from an additional sub-graph matching tool, which enables better structure generation for queried chemical entities, leading to more accurate grounding.
However, grounding agents still exhibit limitations. One key weakness is name-to-structure mapping, where small LLMs like LLaMA 3.1-8B achieve low accuracy (37.9\%). Additionally, agents sometimes match results across different substructures and struggle to filter out irrelevant grounding results based on context. Figures \ref{fig:comparedAgent} and \ref{fig:IrrelevantAndWrongResult} visualizes these issues. Figure \ref{fig:comparedAgent} shows the grounding result for \textit{benzene}. While GPT-4o identifies correctly that \textit{benzene} is a six-membered substructure, its atom indices are scattered across multiple substructures. The grounding agent provides more accurate results but fails to fit the constraints that \textit{Fluorine} substituents described in the caption. Another example in Figure \ref{fig:IrrelevantAndWrongResult} is to show a drawback of using subgraph retrieval technique for RSL where locations are overlapped generated for a chain and are scattered across different substructures.
% , such as \textit{(8, 9, 10, 11, 12, 13)} vs. \textit{(9, 10, 11, 12, 13, 14)}.

\subsection{Can Ground Help Downstream Tasks?}

We conducted experiments to evaluate the impact of molecular grounding on molecular captioning and classification.
For molecular captioning, we incorporate grounding results generated by the grounding agent (GPT4-based) as additional input. As shown in Table \ref{Tab:perfOnCaptioning}, this additional information improved performance across all BLEU metrics. 
\begin{table}[t]
\small
\centering
% \renewcommand{\arraystretch}{1.0}
% \resizebox{0.95\linewidth}{!}{
\newcolumntype{C}[1]{>{\centering\arraybackslash}p{#1}}

\begin{tabular}{C{2cm}|C{0.9cm}|C{0.9cm}|C{0.9cm}|C{0.9cm}}
\bottomrule

\textbf{Model}            & \textbf{BLEU1}  & \textbf{BLEU2}  & \textbf{BLEU3} & \textbf{BLEU4} \\ \hline
     GPT4o           & 30.861 & 15.683 & 8.032 & 4.199 \\ 
                            + Grounding & \textbf{31.178} & \textbf{16.698} & \textbf{9.002} & \textbf{5.004} \\ 
                             % \hline
% \multirow{2}{*}{LLaMA  3.1-8B} & SMILES           &        &        &       &       \\
%                              & SMILES+Grounding &        &        &       &      
% \\ 
\toprule   

\end{tabular}
% }
\caption{Performance on Molecular Captioning}
\label{Tab:perfOnCaptioning}
% \vspace{-0.15in}
\end{table}

\begin{table}[t]
\small
\centering
\newcolumntype{C}[1]{>{\centering\arraybackslash}p{#1}}

% \resizebox{1\linewidth}{!}{
% {C{2.5cm}|C{2.6cm}|C{1.2cm}|C{1.2cm|C{1.2cm}|C{1.2cm}
\begin{tabular}{C{1.6cm}|C{0.7cm}|C{0.7cm}|C{0.7cm}|C{0.7cm}|C{0.7cm}}
\bottomrule

% \textbf{Model}            & \textbf{Aim. $\uparrow$}  & \textbf{Cov. $\uparrow$}  & \textbf{Acc. $\uparrow$} & \textbf{Abs.T $\uparrow$} &\textbf{Abs.F $\downarrow$} &

\textbf{Model}            & \textbf{Aim.}  & \textbf{Cov.}  & \textbf{Acc.} & \textbf{Abs.T } &\textbf{Abs.F} \\

\hline
 ATC-CNN          & 67.86 & 66.65 & 65.04 & 60.65 & \textbf{3.83}\\ 
 + Grounding & \textbf{70.15} & \textbf{71.71} &  \textbf{67.81} & \textbf{61.49} & 4.18\\    
\toprule   

\end{tabular}
% }
\caption{Performance Comparison on Molecular Classification Task. Note that lower Abs.F is better.}
\label{Tab:perfOnATC}
% \vspace{-0.2in}
\end{table}

For molecular classification, we investigated the effectiveness of integrating molecular substructure information into ATC classification (Anatomical, Therapeutic, Chemical). We use ATC-CNN~\cite{cao2022identifying} as the baseline and conduct the experiments on ATC-SMILES~\cite{cao2022identifying} dataset with the resulting substructures.
% we extracted substructure SMILES for each molecule and generated embeddings using an RNN~\cite{medsker2001recurrent}. These embeddings were fused with the original embeddings before classification. 
As shown in Table \ref{Tab:perfOnATC}, incorporating substructure information led to significant performance gains across almost all evaluation metrics, including aiming (+3.37\%), coverage (+7.59\%), accuracy (+4.26\%), and absolute true (+1.38\%), demonstrating that molecular grounding enhances drug classification.

\section{Conclusion}

This paper has introduced a molecular grounding benchmark to enhance the referential aspect of molecular understanding.
MolGround, with 117k QA pairs across five subtasks, is the largest molecular QA benchmark. Our evaluation shows that existing LLMs struggle with these tasks, with SFT and ICL yielding minor improvements.
Our grounding prototype outperforms existing models and enhances downstream tasks like captioning and ATC classification.

\bibliography{custom}

\appendix

\section{Appendix}
\label{sec:appendix}

\subsection{Molecular Captioning Prompt Template}
\label{Appendix_prompt}
\begin{tcolorbox}[boxrule=0pt, frame empty]
    $f_{general}(IUPAC, SMILES)$: Given a molecular IUPAC name and its SMILES, your task is to provide a detailed description, including Basic Structure, Functional Groups, Stereochemistry, Molecular Size and Shape, Physicochemical Properties, Reactivity, Safety and Environmental Impact, etc.
\end{tcolorbox}

\begin{tcolorbox}[boxrule=0pt, frame empty]
$f_{publication}(literature)$: Given a molecular literature, extract the following information from the literature:
1) Physicochemical Properties: includes physicochemical characteristics of the molecule such as hole mobility, molecular weight, solubility, boiling point, melting point, pKa value (acid dissociation constant), and logP (lipophilicity);
2)Safety Information: Provides information regarding the safety of the molecule, such as its toxicity, carcinogenic, teratogenic, or mutagenic properties.
3) Application Areas: Provides an overview of the applications of the molecule.
4) Spectroscopic Properties: include spectroscopic data of the molecule, such as UV-visible absorption spectrum, infrared spectrum, nuclear magnetic resonance spectrum, and mass spectrometry data
\end{tcolorbox}

\begin{tcolorbox}[boxrule=0pt, frame empty]
$f_{specific}(Structure Image, SMILES)$: Given a molecular structure image and SMILES, generate a detailed molecular description (within 100 words) focusing number of rings, their types, and associated properties.
\end{tcolorbox}

\begin{tcolorbox}[boxrule=0pt, frame empty]
$f_{summarize}(f_{general},f_{paper}, f_{specific})$: Given a molecular structure image, SMILES, IUPAC and three initial descriptions, summarize them and generate a molecular description focusing on basic structure, how substructures connect, and outlining their properties. following the example provided below.
$Examples$
\end{tcolorbox}

\subsection{Evaluation Metrics}
\label{Appendex_evaluation_metric}
% The grounding of a subscture is represented as a tuple. 
%

\begin{figure*}[t]  
\centering  
  \centering  
  \includegraphics[width=0.9\textwidth]{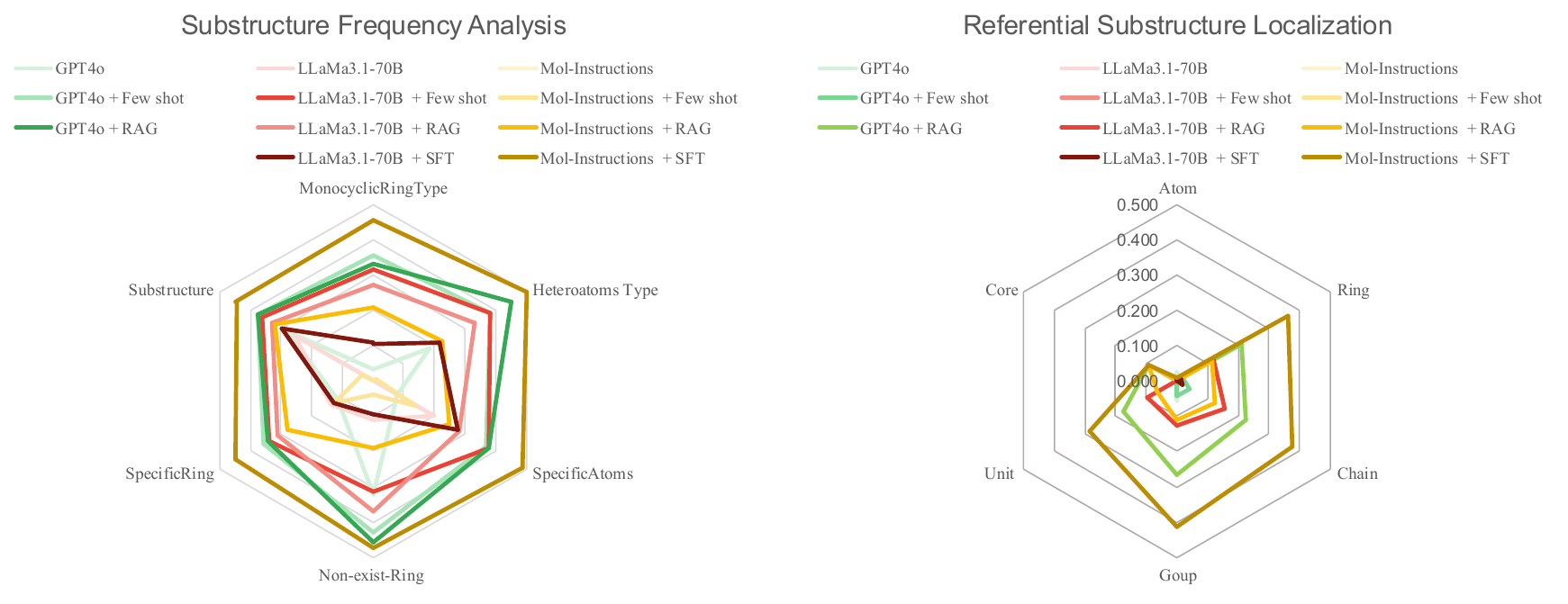}  
    \caption{Performance on fine-grained types and structures.}   
    \label{fig:fine-grained_result}
\end{figure*}

% The grounding of a subscture is represented as a tuple. 
%

For CNER, we report the F1-score of the multi-entity prediction and ground truth. 
For BNSM, SRG and SFA, we report accuracy.
For RSL, we perform both substructure- and molecular-levels evaluation and report F1-score, IoU, accuracy and the substructure coverage. 
For the substructure-level evaluation, we evaluate the ground performance of each instance of each substructure one by one. 
As a substructure could have multiple instances in a molecule, we perform Hungarian matching to find the optical matches and evaluate on the best possible matches. Specifically, we compute the node IoU between grounding prediction and its ground truth of a substructure (i.e., local IoU $IoU_{l}$) and use it as the Hungarian cost function.
We report precision, recall, and F1-score ($F1_{l}$) with $IoU_{l} = \{1, 0.7, 0.5\}$, where $IoU_{l}=1$ represents an exact match between the ground truth and the prediction, and $IoU_{l}=0.7$ (or 0.5) indicates 70\% (or 50\%) of node coverage. 
For the molecule-level evaluation, we treat all the predictions of a substructure as a whole and compare it with the ground truth annotation. Specifically, the molecule is seen as a graph where atom as node and their bonds as edge, and the grounding task is a node binary classification task. 
Specifically, the nodes belonging to the mentioning substructures should be highlighted (i.e., label=1). Otherwise, they should have the label of 0. 
Assuming that the ground truth label for a substructure in the molecule with $m$ atoms is $y = [y_1,...,y_m]$ and the predicted node classification as $\hat{y} = [\hat{y}_1,...,\hat{y}_m]$, we compute the average accuracy of the node classification as the global evaluation metric as:
\begin{equation}
     Acc_{g} = \frac{\#correctPrediction}{\#atoms} = \frac{
     |\hat{y}_i=y_i|
     }{m}
\end{equation}
We also compute the IoU of the substructure $S=\{a_i|y_i=1\}$ and the predicted highlight nodes $P=\{a_i|\hat{y}_i=1\}$ as another global metric:
\begin{equation}
     IoU_{g} = \frac{S \cap P}{S \cup P}
\end{equation}
Besides, for the multiple substructure grounding task, we also report the substructure coverage rate $Cov_{s}$.

\subsection{Fine-grained Performance}
We also visualize the comparison on fine-grained level in Figure \ref{fig:fine-grained_result}.
The ICL and SFT significantly improve the SFA results on all types of counting tasks. 
The best-performing model is Mol-Instructions with SFT, which has over 0.9\% on all dimensions.
For the RSL, all models perform poorly at the atom and core grounding.

\end{document}